\documentclass[conference]{IEEEtran}
\IEEEoverridecommandlockouts
\usepackage{cite}
\usepackage{amsmath,amssymb,amsfonts}
\usepackage{algorithmic}
\usepackage{graphicx}
\usepackage{textcomp}
\usepackage{xcolor}
\usepackage{booktabs}
\usepackage{multirow}
\def\BibTeX{{\rm B\kern-.05em{\sc i\kern-.025em b}\kern-.08em
    T\kern-.1667em\lower.7ex\hbox{E}\kern-.125emX}}
\begin{document}

\title{SmoothGuard: Defending Multimodal Large Language Models with Noise Perturbation and Clustering Aggregation\\

\thanks{\textsuperscript{*}Equal contribution.
\textsuperscript{‡}Corresponding author.}
}

\author{
\IEEEauthorblockN{
Guangzhi Su\textsuperscript{\rm 1,*},
Shuchang Huang\textsuperscript{\rm 2,*},
Yutong Ke\textsuperscript{\rm 1},
Zhuohang Liu\textsuperscript{\rm 1},
Long Qian\textsuperscript{\rm 1},
Kaizhu Huang\textsuperscript{\rm 1,\dag}
}
\IEEEauthorblockA{\textsuperscript{\rm 1} Division of Natural and Applied Sciences, Duke Kunshan University, Suzhou, China\\
{guangzhi.su, yutong.ke, zhuohang.liu, long.qian, kaizhu.huang}@dukekunshan.edu.cn}
\IEEEauthorblockA{\textsuperscript{\rm 2} Independent Researcher\\
shuchanghuang2001@gmail.com}
}

\maketitle

\begin{abstract}
Multimodal large language models (MLLMs) have achieved impressive performance across diverse tasks by jointly reasoning over textual and visual inputs. Despite their success, these models remain highly vulnerable to adversarial manipulations, raising concerns about their safety and reliability in deployment. In this work, we first generalize an approach for generating adversarial images within the HuggingFace ecosystem and then introduce SmoothGuard, a lightweight and model-agnostic defense framework that enhances the robustness of MLLMs through randomized noise injection and clustering-based prediction aggregation. Our method perturbs continuous modalities (e.g., images and audio) with Gaussian noise, generates multiple candidate outputs, and applies embedding-based clustering to filter out adversarially influenced predictions. The final answer is selected from the majority cluster, ensuring stable responses even under malicious perturbations. Extensive experiments on POPE, LLaVA-Bench (In-the-Wild), and MM-SafetyBench demonstrate that SmoothGuard improves resilience to adversarial attacks while maintaining competitive utility. Ablation studies further identify an optimal noise range (0.1–0.2) that balances robustness and utility. These findings establish SmoothGuard as a practical step toward secure and reliable multimodal reasoning, with future work extending the framework to audio-focused benchmarks to validate its generality across modalities. Code will be publicly available soon on Github.
\end{abstract}


\section{Introduction}
\label{Introduction}
In recent years, multi-modal large language models (MLLMs) \cite{pi2024mllm, chen2024mllm, zhang2024differential, bai2025qwen2,ye2024mplug} have surged in popularity, driven by their capacity to assimilate and analyze data across text and visual modalities. These models find applications in diverse domains, including content generation, question answering, and visual-language reasoning. Their sophisticated ability to synchronize multimodal inputs enables them to deliver coherent and precise responses, showcasing significant advancements in artificial intelligence technologies\cite{gao2024cantor}.

Despite their impressive capabilities, MLLMs are not immune to adversarial attacks that hurt their functionality and safety\cite{pi2024mllm, liu2024safety}. Adversarial inputs were designed deliberately by data manipulations, which pose a serious threat as they can induce significant misclassifications and trigger potentially harmful responses. Current adversarial attacks on MLLMs are mostly targeted on continuous modality like image and audio by adding designed noise to get close and cross the decision boundary. These threats underscore the urgent need for robust defenses to facilitate the safe deployment of MLLMs in practical scenarios.

To this end, various strategies \cite{chen2024bathe, du2024vlmguard}have been proposed to provide defenses for both large language models (LLMs) and MLLMs. Innovations such as SMOOTHLLM and patched visual prompt defenses represent notable attempts to safeguard these models. However, these measures often fall short when confronted with sophisticated, multi-modal adversarial strategies. What's more, in resource constrained circumstances, finetuning a model with a large designed dataset might be infeasible, which calls for a method that can neutralize the malicious prompt during inference stage, and a plug-and-play method that can be applied to different models directly.

To address these issues, we propose a defense strategy based on randomized noise injection and semantic clustering method for selecting the optimal answer. Unlike detector-based or heuristic defenses, our method leverages multiple perturbed inputs and aggregates their predictions through clustering, filtering out adversarially influenced responses while preserving stable ones. This design provides robustness against subtle, hard-to-detect attacks without requiring model retraining or architectural modifications. Initial experiments on vision–language benchmarks show that the method reduces adversarial risk while retaining competitive utility, with ablation studies identifying a practical range of noise levels that balances these objectives. Our work thus contributes a lightweight, model-agnostic approach to enhancing the reliability of MLLMs in adversarial settings.

This paper is structured as follows: Section 2 reviews related work in the realm of adversarial attacks and defenses for MLLMs. Section 3 explains our proposed methodology in detail. Section 4 describes our experimental framework and discusses the results. Section 5 concludes with a reflection on our findings and suggests directions for future research.
\begin{figure*}[t]
  \centering
  \includegraphics[width=0.7\textwidth]{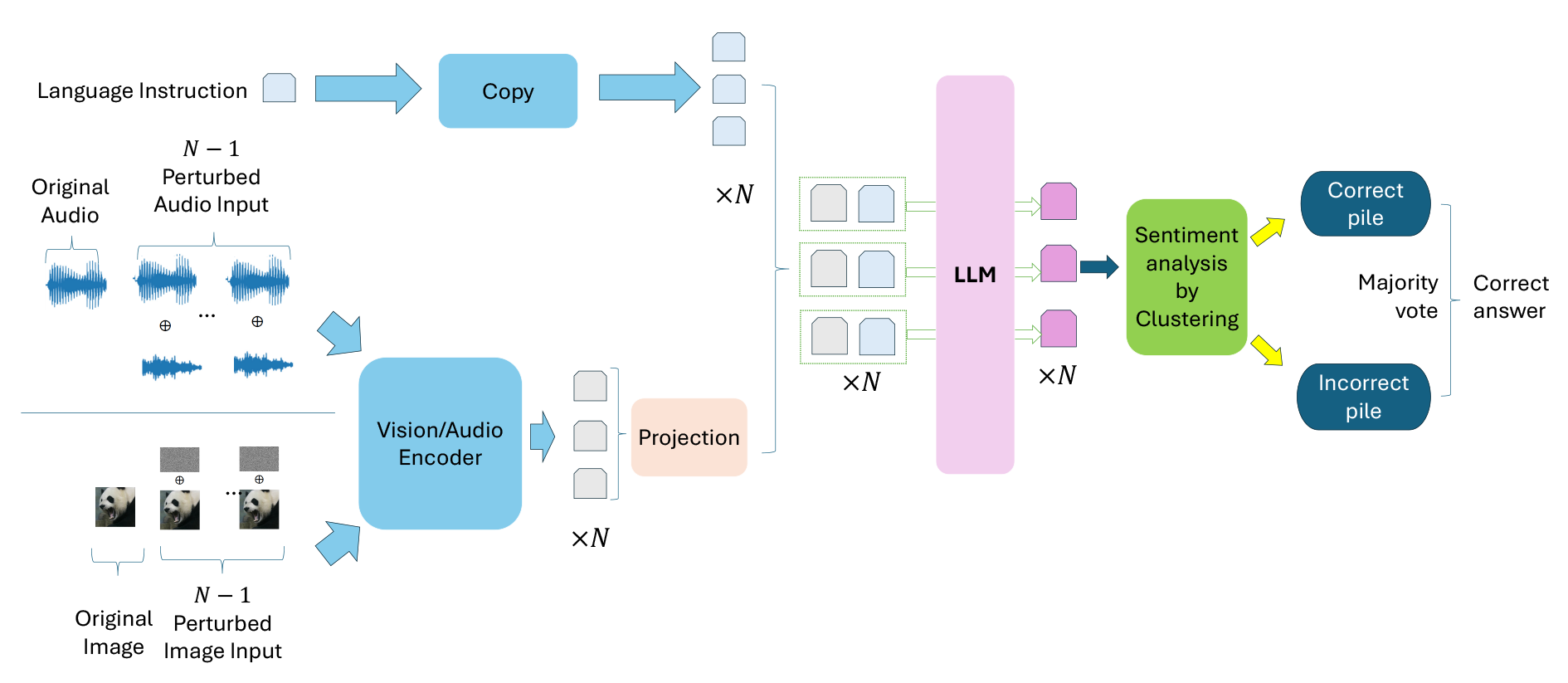} 
  \caption{Overall pipeline of our randomized smoothing defense for MLLMs. 
  Perturbed audio and image inputs are processed through the encoder and projection, 
  producing $N$ responses via the language model. Outputs are clustered and sentiment-aware majority voting 
  is applied to select the final robust answer.}
  \label{fig:framework}
\end{figure*}
\section{Related Work}

\subsection{Adversarial Attacks on Multimodal Models}

The attack surface of multimodal MLLMs has expanded beyond unimodal paradigms \cite{zhou2024mathattack, goodfellow2014explaining, szegedy2013intriguing}, with cross-modal vulnerability propagation emerging as a critical threat vector. Crucially, multimodal jailbreaking exploits safety mechanisms via perturbation vectors \cite{wang2024llms, qi2024visual} and in-context learning backdoors \cite{wen2025investigating}, creating attack surfaces that transcend traditional unimodal paradigms. Contemporary research demonstrates that imperceptible perturbations can induce cascading failures across modalities: multimodal systems exhibit significant degradation in safety performance under semantically constrained adversarial attacks, where signal-level distortions propagate vulnerabilities between audio and textual domains \cite{zhao2023evaluating}, while audio-language systems exhibit alarming susceptibility to jailbreak attacks. Chen et al. \cite{chen2025audiojailbreak} achieve asynchrony, universality, stealthiness, and over-the-air robustness at the same time via adversarial suffix injections to benign audio carriers. Ma et al. \cite{ma2025audio} compromise SpeechGPT through discrete token manipulation in HuBERT embeddings. These vulnerabilities are systematically quantified in the AJailBench benchmark \cite{song2025audio}, reporting a high attack success rate (ASR) across seven state-of-the-art models through Bayesian-optimized perturbations targeting ten prohibited categories.

\subsection{Defensive Techniques in Single and Multimodal Models}

Defensive strategies for single-modality models provide a foundation for protecting MLLMs.\cite{qian2022survey,lyu2015unified} Unimodal foundations include randomized smoothing for certified vision robustness \cite{cohen2019certified}, ensemble methods preserving text integrity under synonym substitution attacks \cite{tramer2017ensemble}, and Time-Domain Noise Flooding (TDNF), reducing audio jailbreak ASR \cite{peri2024speechguard}. For multimodal scenarios, SPIRIT's neuron-level patching achieves a 99\% defense success rate (DSR) through activation modification \cite{djanibekov2025spirit}, while hybrid purification frameworks integrate input transformation with feature sanitization  \cite{sun2024safeguarding, wen2025investigating}. Persistent limitations remain in mitigating real-time audio adversarial inputs  \cite{chen2025audiojailbreak} and certifying robustness against cross-modal perturbation transfer, particularly for safety-critical deployments requiring low-latency responses.

Our method extends these defense strategies by applying random sampling and aggregation techniques to image and audio inputs, improving the adversarial robustness of MLLMs. While previous work has focused on single-modality defenses or simplified approaches to multimodal models, our contribution offers a unified defense that covers both modalities. A novel clustering-based sentiment analysis method then enables extension to a much wider range of adversarial attack tasks.

\section{Proposed Method}

\subsection{Overview}
We propose a robust defense method for MLLMs as shown in Fig.~\ref{fig:framework}, which involves injecting noise to continuous modalities (e.g. image/audio) to safeguard against adversarial attacks. Our strategy is to apply a diverse array of perturbations to the image and then aggregate the model outputs to generate a consolidated, robust prediction. This approach is designed to defend against sophisticated adversarial attacks by sampling a variety of perturbations and clustering the resulting outputs. This clustering helps to filter out outputs that are likely to have been influenced by adversarial inputs, thus ensuring a more reliable and secure response.

Additionally, we employ sentiment analysis based majority voting as a key component of our defense mechanism. This method allows us to evaluate the consensus among the different perturbed outputs, effectively defending against both task-specific and jailbreaking attacks. By doing so, we ensure that our model's responses remain dependable even under malicious conditions.

The effectiveness of our approach is further enhanced by integrating feedback loops from the sentiment analysis process, which help to continually refine the perturbation and clustering techniques. This adaptive mechanism not only improves the resilience of our MLLMs over time but also maintains high accuracy and performance across a range of tasks. By providing a robust framework for handling adversarial inputs in multimodal scenarios, our method stands as a significant advancement in the field of AI security.

\subsection{Random Sampling of Perturbations}
Let $x_{\text{img}}$, $x_{\text{text}}$, and $x_{\text{audio}}$ represent the clean image, text, and audio inputs, respectively. To improve robustness, we generate perturbed versions of the continuous modalities by introducing random noise. We keep text unchanged to preserve semantic integrity after observing a severe performance drop with changed text.

For the image input, Gaussian perturbations with variance $\sigma_{\text{img}}^2$ are applied:
\[
\delta_{\text{img}} \sim \mathcal{N}(0, \sigma_{\text{img}}^2 I), \quad 
\tilde{x}_{\text{img}} = x_{\text{img}} + \delta_{\text{img}}. \tag{1}
\]

For the audio input, we apply a similar perturbation at the waveform or spectrogram level:
\[
\delta_{\text{audio}} \sim \mathcal{N}(0, \sigma_{\text{audio}}^2 I), \quad 
\tilde{x}_{\text{audio}} = x_{\text{audio}} + \delta_{\text{audio}}. \tag{2}
\]

The text input is used in its original form:
\[
\tilde{x}_{\text{text}} = x_{\text{text}}. \tag{3}
\]

Thus, for each sampling step, we obtain randomized variations of the image and audio inputs, while maintaining consistent textual guidance.

\subsection{Aggregation of Predictions via Clustering}
Once we obtain perturbed versions of the image and audio inputs, we also include one unperturbed copy of the original input to preserve utility. Each input pair $(\tilde{x}_{\text{img}}^{(i)}, \tilde{x}_{\text{audio}}^{(i)}, x_{\text{text}})$ is passed through the MLLM to generate a prediction:
\[
y_{\text{pred}}^{(i)} = f(\tilde{x}_{\text{img}}^{(i)}, \tilde{x}_{\text{audio}}^{(i)}, x_{\text{text}})
\tag{4}
\]
where $i \in \{1, 2, \dots, N\}$ indexes the perturbed samples, and an additional sample corresponds to the original unperturbed input.

\begin{figure}[t]
  \centering
  \includegraphics[width=\linewidth]{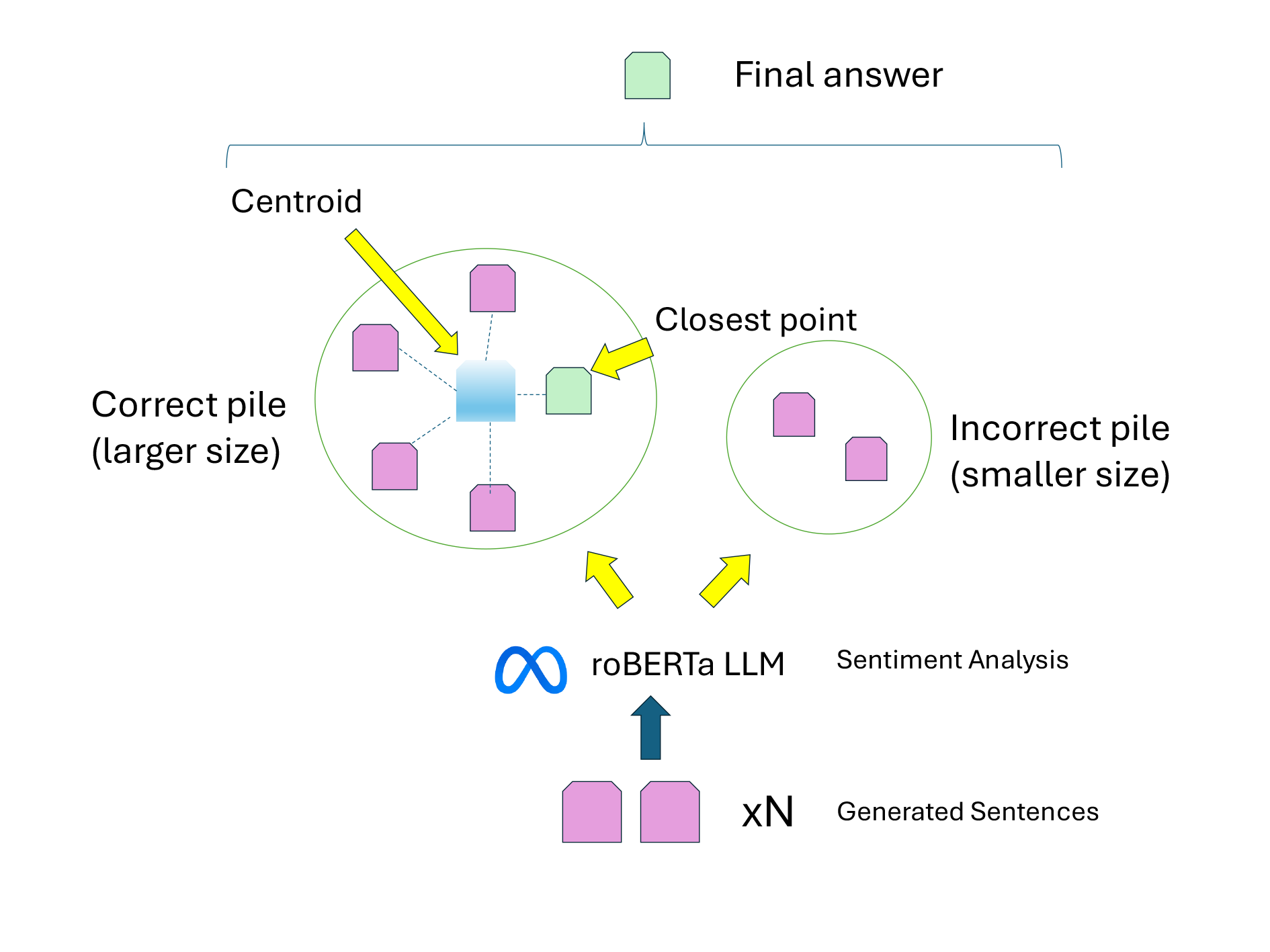}
  \caption{Clustering-based aggregation of generated sentences.}
  \label{fig:clustering-agg}
\end{figure}

As shwon in Fig.~\ref{fig:clustering-agg}, to aggregate predictions, we embed the outputs using a \textit{RoBERTa-base} encoder, producing a fixed-dimensional representation $\mathbf{e}_i$ for each prediction.\cite{liu2019roberta} We then apply \textit{k-means clustering} with $k=2$:
\[
\text{clusters} = \text{k-means}(\{\mathbf{e}_1, \mathbf{e}_2, \dots, \mathbf{e}_N\}, k=2).
\tag{5}
\]

Intuitively, robust models tend to embed semantically-equivalent inputs into tight, well-separated clusters, while adversarial perturbations tend to push corrupted inputs away from the clean cluster into different regions in the representation space. This effect has been observed in other researchers' papers related to the “clustering effect” in robust networks)\cite{bai2021clustering}. In practice, we often observe a dominant clean cluster covering most of the perturbed samples, with the remainder forming a distinct minority cluster corresponding to adversarial deviations. This motivates our use of k = 2 for clustering: one larger cluster for clean and stable predictions, and the other for adversarial ones.

We then select the larger cluster for next step process. The centroid of this cluster is computed, and the prediction that is the closest to the centroid was chosen as the representative answer for stability:
\[
\hat{y} = \arg\max_{\mathbf{e} \in \text{cluster}} \cos(\mathbf{e}, \text{centroid}),
\tag{6}
\]
where $\cos(\mathbf{e}, \text{centroid})$ denotes the cosine similarity.

Including the unperturbed input ensures that, under benign conditions, the final prediction remains close to the model’s original utility. In the presence of adversarial attacks, the use of multiple perturbed copies prevents a single corrupted prediction from dominating the outcome, thereby stabilizing the defense.

\section{Experiment}

\subsection{Experimental Setup}
All experiments were implemented using the HuggingFace pipeline, which allows for straightforward adaptation to different MLLMs. Our study focuses primarily on the Qwen2.5-VL-7B and LLaVA-1.5-7B models, with experiments organized into two parts: robustness evaluation and utility evaluation. The robustness evaluation measures defense performance under adversarially perturbed inputs, while the utility evaluation examines performance under benign conditions. For the latter, we consider two representative tasks: (1) multiple-choice question answering, and (2) free-form sentence-based question answering. Finally, we conduct an ablation study to identify the optimal range of noise levels, aiming to balance robustness improvements with minimal utility degradation.

\begin{table*}[t]
\centering
\setlength{\tabcolsep}{6pt}
\renewcommand{\arraystretch}{1.15}
\caption{\textbf{Evaluation on MM-SafetyBench}: We report the ASR$\downarrow$ across seven prohibited categories, using LlamaGuard-7B as the jailbreak classifier. Bold indicates the lowest ASR per column. SmoothGuard uses randomized smoothing ($\sigma=0.1$; 9 noisy + 1 original, majority vote).}
\label{tab:mmbench-top7}
\resizebox{\textwidth}{!}{%
\begin{tabular}{l l c c c c c c c c}
\toprule
Model & Method & Illegal Activity & Hate Speech & Malware Generation & Physical Harm & Economic Harm & Fraud & Sex & Avg$\downarrow$ \\
\midrule
\multirow{2}{*}{Qwen2.5-VL-7B}
& Original     & 0.0412 & \textbf{0.0000} & 0.0909 & 0.0417 & \textbf{0.0000} & 0.0130 & 0.0550 & 0.0345 \\
& SmoothGuard  & \textbf{0.0000} & \textbf{0.0000} & \textbf{0.0682} & \textbf{0.0139} & \textbf{0.0000} & \textbf{0.0000} & \textbf{0.0459} & \textbf{0.0183} \\
\midrule
\multirow{2}{*}{LLaVA-1.5-7B}
& Original     & 0.5876 & 0.2945 & 0.3409 & 0.5208 & 0.0984 & 0.5130 & 0.5413 & 0.4138 \\
& SmoothGuard  & \textbf{0.2887} & \textbf{0.1411} & \textbf{0.2727} & \textbf{0.2986} & \textbf{0.0164} & \textbf{0.1688} & \textbf{0.4587} & \textbf{0.2350} \\
\bottomrule
\end{tabular}}
\end{table*}

\subsection{Robustness Analysis}
\textbf{Benchmark and attack.} For robustness analysis, we adopt MM-SafetyBench as the benchmark.\cite{liu2024mm} This benchmark comprises 13 scenarios and contains over 5,000 text–image pairs. In our evaluation, however, we replace the dataset images with a single universal adversarial image trained in advance, as this aligns with our target adversarial attack method and is generally more effective than standard attacks. Our experiments build upon the adversarial image attack strategy proposed by \cite{qi2024visual}, but we observed that the transferability of the perturbed adversarial image is limited. To address this, we modify the original pipeline and generalize the implementation such that any MLLM hosted on the HuggingFace platform can be paired with a corresponding adversarial image by simply specifying the model name.

\textbf{Protocol and Metric.} We adopt randomized noise injection with additive Gaussian noise and set the default noise level to $\sigma$ = 0.1. At inference, we perform 10 stochastic evaluations per item and aggregate predictions via majority vote based on our method. To determine whether the unsafe prompt still successfully jailbreaks the model, we use an external safety classifier, LlamaGuard-7B. \cite{inan2023llama} A sample is counted as an attack success if LlamaGuard flags the model’s response as violating the safety policy for the given unsafe prompt. We focus on the ASR and report per-category values.

\textbf{Results}. Table~\ref{tab:mmbench-top7} summarizes ASR across seven prohibited categories on MM-SafetyBench. For Qwen2.5-VL-7B, SmoothGuard consistently lowers ASR relative to the original model, nearly eliminating successful attacks in categories such as Illegal Activity and Physical Harm. Similar improvements are observed on LLaVA-1.5-7B, where average ASR drops substantially across all categories. Beyond the raw reductions, these results highlight that a lightweight defense can generalize across architectures with very different baseline vulnerabilities. Taken together, the findings confirm SmoothGuard’s effectiveness as a universal defense for multimodal safety.

\subsection{Utility Analysis}
For utility analysis, we conducted experiments in two settings: multiple-choice question answering and sentence-based question answering. Evaluations were performed on both the Qwen2.5-VL-7B and LLaVA-1.5-7B models, comparing performance under Gaussian noise ($\sigma$ = 0.1) against the no-noise baseline.

\textbf{Multiple-choice Question.} For evaluating multiple-choice performance, we employ the POPE benchmark, a multi-modal dataset comprising 9,000 vision–language pairs. POPE is organized into three categories—\emph{Adversarial}, \emph{Popular}, and \emph{Random}—which are specifically designed to assess hallucination tendencies in vision–language models. These categories provide a rigorous framework for revealing the robustness and reliability of model predictions. To ensure a comprehensive evaluation, we report standard classification metrics, including Accuracy, Precision, Recall, and F1 score.

\begin{figure}[t]
    \centering
    \includegraphics[width=0.9\linewidth]{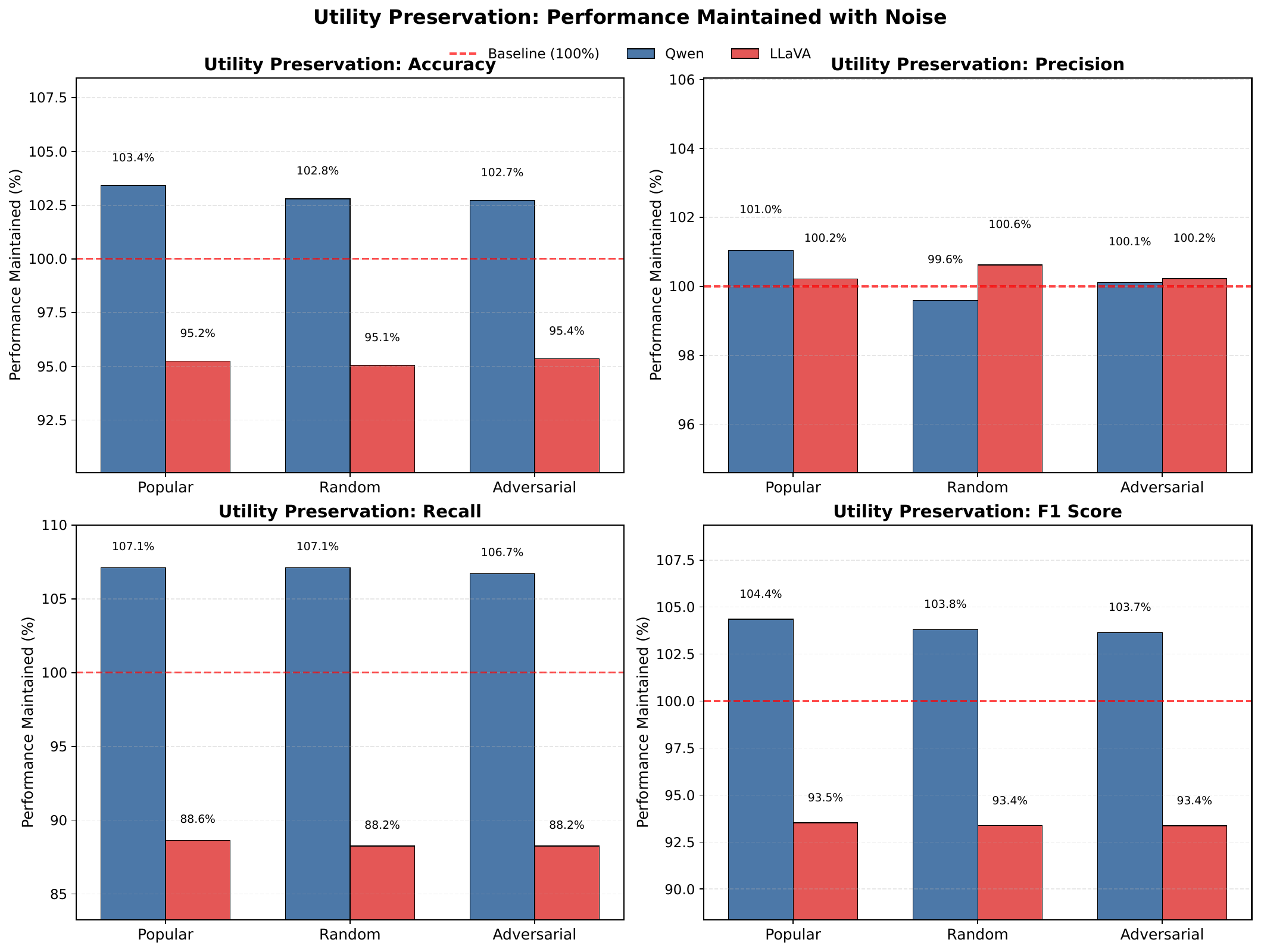}
    \caption{Utility preservation of Qwen and LLaVA with randomized smoothing noise across different categories. Performance is normalized to the baseline (100\%), showing that Qwen improves utility while LLaVA maintains competitive results.}
    \label{fig:utility-preservation}
\end{figure}

As shown in Fig.~\ref{fig:utility-preservation}, our results demonstrate that the introduction of randomized smoothing preserves the utility of both models across different categories. For Qwen, smoothing leads to consistent improvements, with accuracy, precision, recall, and F1 score all exceeding the baseline (100\%) across the \emph{Popular}, \emph{Random}, and \emph{Adversarial} settings. In contrast, LLaVA exhibits only a modest decline, but performance remains close to the baseline, confirming that utility is largely maintained even with noise injection. These results indicate that randomized smoothing enhances robustness while ensuring that model utility is not substantially compromised.

Overall, this analysis highlights the practicality of our approach: randomized smoothing provides a reliable defense mechanism that strengthens robustness without sacrificing the effectiveness of MLLMs in utility-driven tasks.

\textbf{Sentence Question Answering.} For sentence-level question answering, we adopt the \emph{LLaVA-Bench (In-the-Wild)}, which consists of diverse image–text pairs divided into three categories: \emph{Conv} (short conversational queries), \emph{Detail} (fine-grained descriptions of the image), and \emph{Complex} (multi-step reasoning and more challenging scenarios). We evaluate both models' answers alongside a reference answer produced by GPT. To assess performance, we prompt the latest LLM model to act as a judge, rating the quality of model answers relative to the GPT baseline.

\begin{figure}[htbp]
\centering
\includegraphics[width=0.9\linewidth]{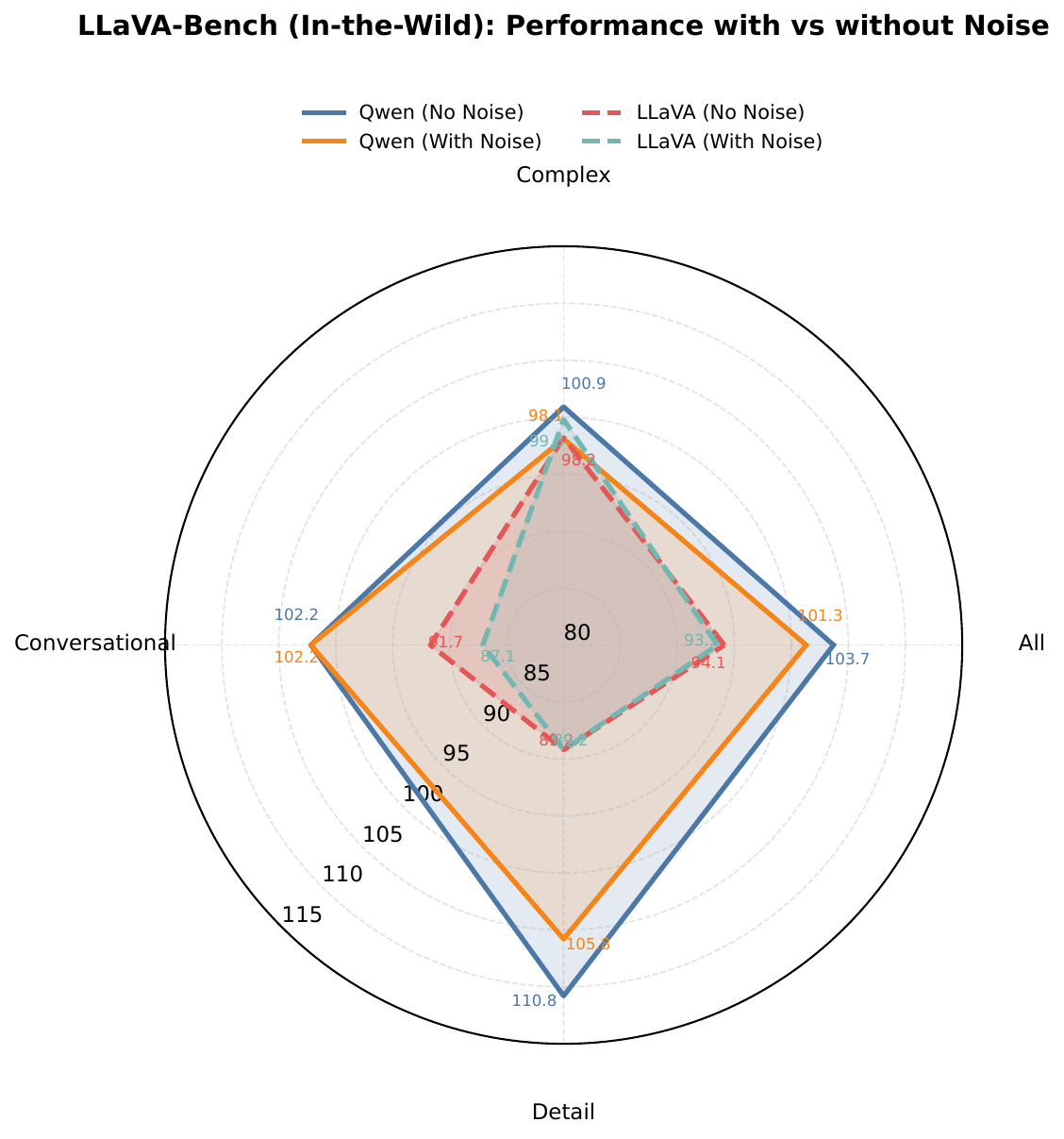}
\caption{LLaVA-Bench (In-the-Wild) relative scores of Qwen and LLaVA with and without randomized smoothing noise, compared against GPT-4 (higher is better).}
\label{fig:wildbench-radar}
\end{figure}

As shown in Fig.~\ref{fig:wildbench-radar}, introducing randomized smoothing leads to only marginal performance differences for both models. For Qwen, the scores with noise remain close to those without noise across all categories, indicating that utility is effectively preserved. For LLaVA, utility is also largely maintained: there is a small drop in the \emph{Conversational} category, a slight improvement in the \emph{Complex} category, and negligible changes in the \emph{Detail} and \emph{All} categories. Overall, these results suggest that randomized smoothing provides robustness benefits while preserving utility, with only limited trade-offs across different sentence-level tasks.

\subsection{Ablation Study}
Before conducting the main experiments, we performed an ablation study to investigate the effect of varying noise levels on model robustness and utility. 

\textbf{Utility} For utility, we ran the Qwen2.5-VL-7B model on a subset of the POPE dataset under the adversarial setting and evaluated its accuracy across different Gaussian noise levels.

\begin{figure}[htbp]
    \centering
    \includegraphics[width=0.8\linewidth]{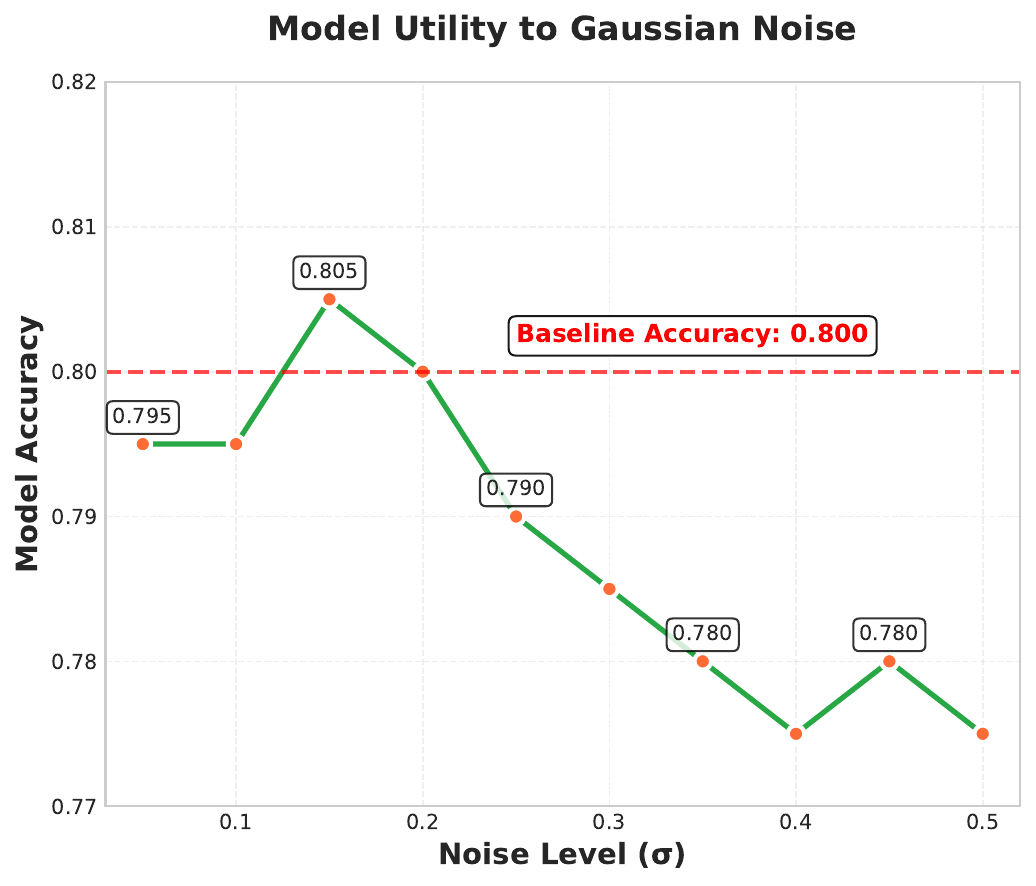}
    \caption{Model Utility to Gaussian noise on the adversarial setting.}
    \label{fig:accuracy-vs-sigma}
\end{figure}

As illustrated in Fig.~\ref{fig:accuracy-vs-sigma}, model performance remains relatively stable across the tested range, but exhibits a clear optimal region around noise levels of 0.1–0.2. Within this interval, accuracy reaches its peak and demonstrates less fluctuation compared to higher noise values. Beyond this range, particularly after 0.3, we observe a gradual decline in accuracy, suggesting that excessive noise begins to distort the input distribution and impair utility.

This trend highlights a critical trade-off: introducing a small degree of noise can improve robustness without significantly sacrificing accuracy, but overly large perturbations could possibly degrade model performance. Based on these observations, we select 0.1–0.2 as the optimal range of noise levels for the subsequent experiments.

\textbf{Robustness}
To assess sensitivity to the noise magnitude, we conduct an ablation on a representative subset of MM-SafetyBench scenarios while keeping the repeated-evaluation protocol fixed. We vary the noise level from 0.05 to 0.50 in increments of 0.05 and report the ASR as the sole metric.

\begin{figure}[htbp]
    \centering
    \includegraphics[width=0.8\linewidth]{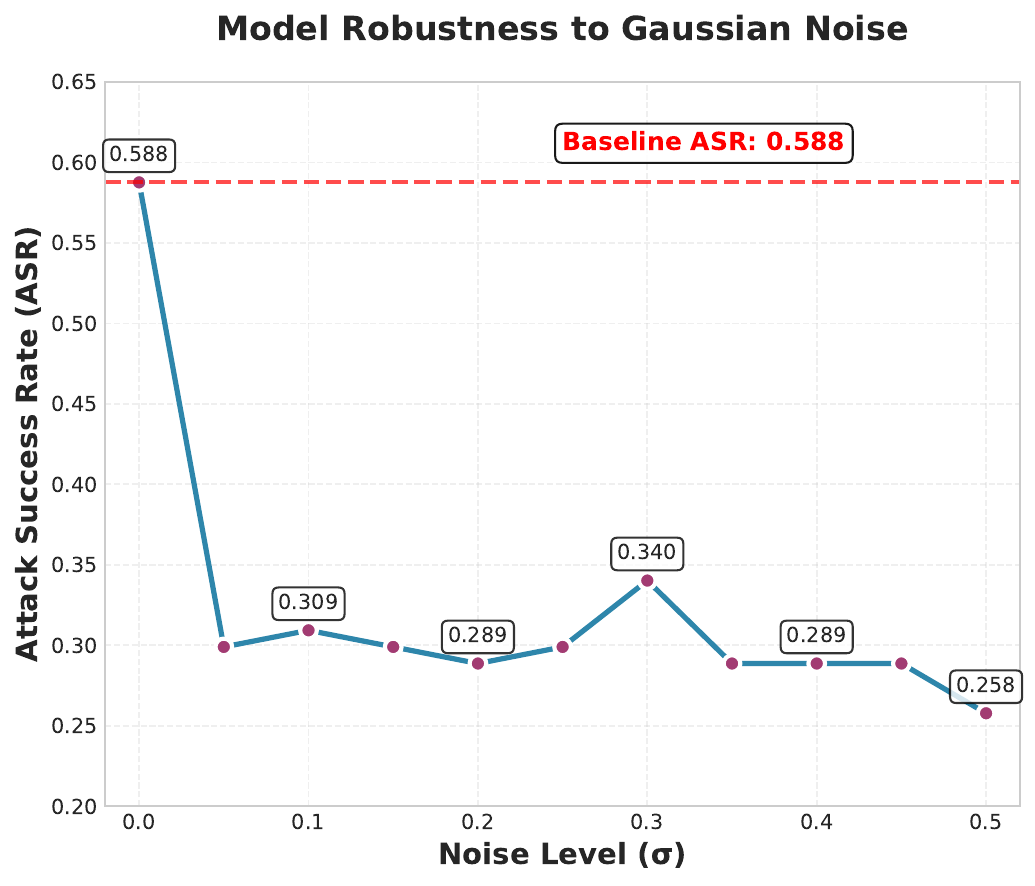}
    \caption{Model Robustness to Gaussian noise on the adversarial setting.}
    \label{fig:robustness-vs-sigma}
\end{figure}

Fig.~\ref{fig:robustness-vs-sigma} shows the change of the ASR. Compared to the baseline ASR of 0.588, even small perturbations substantially improve robustness: ASR drops to 0.309 at $\sigma$ = 0.10 and reaches its lowest value of 0.258 at $\sigma$ = 0.30. Beyond this point, the curve flattens, with no consistent gains at higher noise levels. These results suggest that moderate noise in the range of 0.10–0.30 achieves the most favorable trade-off, effectively reducing adversarial vulnerability while avoiding unnecessary degradation. The 10-pass majority vote further stabilizes predictions, reducing variance across runs.

\section{Conclusion}
In this work, we introduced defense framework for MLLMs by introducing random noise and semantic clustering techniques, targeting adversarial manipulations in continuous modalities such as images and audio. By perturbing inputs with Gaussian noise and aggregating predictions, our method effectively enhances robustness against adversarial attacks while preserving task utility. Experiments on POPE, Bench-in-the-Wild, and MM-SafetyBench demonstrate that randomized smoothing mitigates hallucinations and adversarial vulnerabilities with only marginal performance degradation under benign conditions, achieving results that remain competitive with strong baselines. Ablation studies further reveal that modest noise levels (0.1–0.2) strike the best balance between robustness and utility. These findings highlight randomized smoothing as a lightweight, model-agnostic, and practical defense for securing MLLMs. In future work, we will extend our experiments to the audio modality, further validating the effectiveness of randomized smoothing beyond vision–language tasks.

\bibliographystyle{IEEEtran}
\bibliography{reference}

\end{document}